\documentclass[accepted]{uai2021} % after acceptance, for a revised
\usepackage{silence}
\usepackage[american]{babel}

\usepackage[utf8]{inputenc} % allow utf-8 input
\usepackage[T1]{fontenc}    % use 8-bit T1 fonts
\usepackage{hyperref}       % hyperlinks
\usepackage{url}            % simple URL typesetting
\usepackage{booktabs}       % professional-quality tables
\usepackage{amsfonts}       % blackboard math symbols
\usepackage{nicefrac}       % compact symbols for 1/2, etc.
\usepackage{microtype}      % microtypography

\usepackage[ruled,vlined]{algorithm2e}
\usepackage{multirow}
\usepackage{amssymb}
\usepackage{amsmath}
\usepackage{bm}
\usepackage{pifont}
\usepackage[table]{xcolor}
\usepackage{times}
\usepackage{graphicx} % more modern
\usepackage{subcaption}
\usepackage{wrapfig}
\usepackage{multirow}

\usepackage{hyperref}

\usepackage{natbib} % has a nice set of citation styles and commands
\usepackage{mathtools} % amsmath with fixes and additions
\usepackage{booktabs} % commands to create good-looking tables
\usepackage{tikz} % nice language for creating drawings and diagrams
\renewcommand{\thefootnote}{\fnsymbol{footnote}} 
\newcommand\nnfootnote[1]{%
  \begin{NoHyper}
  \renewcommand\thefootnote{}\footnote{#1}%
  \addtocounter{footnote}{-1}%
  \end{NoHyper}
}

\title{Multi-output Gaussian Processes for Uncertainty-aware Recommender Systems}

\author[1,2]{\href{mailto:Yinchong Yang <yinchong.yang@siemens.com>}{Yinchong Yang}{}}
\author[1,2,3,4,5]{\href{mailto: Florian Buettner <buettner.florian@siemens.com>}{Florian Buettner\footnote{test}}{}}
\affil[1]{%
    Siemens AG
}
\affil[2]{%
    Contributed equally to this work
}

\affil[3]{German Cancer Consortium (DKTK)}
\affil[4]{German Cancer Research Center (DKFZ)}
\affil[5]{Goethe University Frankfurt}
\begin{document}

\maketitle

\begin{abstract} 
Recommender systems are often designed based on a collaborative filtering approach, where user preferences are predicted by modelling interactions between users and items. Many common approaches to solve the collaborative filtering task are based on learning representations of users and items, including simple matrix factorization, Gaussian process latent variable models, and neural-network based embeddings.
While matrix factorization approaches fail to model nonlinear relations, neural networks can potentially capture such complex relations with unprecedented predictive power and are highly scalable. However, neither of them is able to model predictive uncertainties. In contrast, Gaussian Process based models can generate a predictive distribution, but cannot scale to large amounts of data. 
In this manuscript, we propose a novel approach combining the representation learning paradigm of collaborative filtering with multi-output Gaussian processes in a joint framework to generate uncertainty-aware recommendations. We introduce an efficient strategy for model training and inference, resulting in a model that scales to very large and sparse datasets and achieves competitive performance in terms of classical metrics quantifying the reconstruction error. In addition to accurately predicting user preferences, our model also provides meaningful uncertainty estimates about that prediction. 

%main area: rs/cf
%second: rep/GP

% The purpose of this document is to provide both the basic paper template and submission guidelines. Abstracts should be a single paragraph, between 4--6 sentences long, ideally.  Gross violations will trigger corrections at the camera-ready phase.

\end{abstract} 
%\footnotetext{\textsuperscript{\rm * }Work done for Siemens AG}
\nnfootnote{\textsuperscript{\rm * }Work done for Siemens AG}
\section{Introduction}\label{sec:intro} 
Collaborative filtering (CF) provides a powerful solution to recommender systems. 
Recommending a new item to a user is based on the assumption that users demonstrating similar rating or purchasing patterns are interested in similar items. 
A database describing such user-item interactions often takes the form of a matrix, where each entry describes the interaction between one user and one item.  
The overall rating or purchasing pattern of a user can therefore be described by the corresponding row in such a matrix. 
However, since there are typically large numbers of users and items in the database, and each user is usually only interested in a small subset of items, this user-item matrix is often large and sparse. 
It is therefore inefficient to define the similarity between users in the high dimensional feature space defined by all items. 
Instead, it is more advantageous to derive abstract feature vectors that represent users and items, which inspired a large variety of low-rank matrix decomposition models such as non-negative matrix decomposition \citep{zhang2006learning}, biased matrix decomposition \citep{koren2009matrix} and non-parametric decomposition \citep{yu2009fast}. These methods aim at learning low dimensional representations for all users and items, allowing for the prediction of the unobserved interaction between a new pair of user and item. 

Most of these state-of-the-art representation learning methods in form of matrix decomposition, however, focus on point estimates and do not quantify any predictive uncertainty. This limits the applicability of these methods in safety-critical situations. 
For instance, the recommendation of a treatment in the context of clinical decision support needs to be interpreted with much more caution than the recommendation of movies, and transparent uncertainty estimates are crucial to assess the quality of such recommendation. Similarly, when recommender systems are used to detect novel drug-target interactions to facilitate new drug discoveries, it is crucial for such systems to be uncertainty aware, since large costs are associated with studies required for target validation. Other applications could arise in the context of multi-armed bandits for exploration and exploitation during recommendation. 
Furthermore, missing ratings in collaborative filtering tasks are usually not completely random \citep{schnabel2016recommendations, marlin2012collaborative}, and good performance in terms of RMSE and MAE does not always promise good recommendations\citep{cremonesi2010performance}. In this context, the predictive uncertainty serves as a practical way to make this bias in the training data transparent in a principled manner. The model can communicate the uncertainty of a prediction to the user, such that the user knows which individual predictions are likely to translate into good recommendations.

Gaussian processes (GP) are a well known class of models that generates prediction uncertainty alongside point estimates via its kernel function, which measures the similarity between data samples. 
While Gaussian processes are primarily used to solve a wide range of supervised modelling tasks, they have also been applied to perform probabilistic matrix decomposition with a focus on dimensionality reduction, via the Gaussian Process Latent Variable Model (GPLVM). 
Here, a Gaussian process is used as a prior distribution for a function that maps a low-dimensional latent space to the high-dimensional data matrix. 
The individual components of this mapping function are modelled as independent draws from a Gaussian process, assuming all dimensions of the data matrix are independent. 
This independence assumption is problematic in the context of collaborative filtering, where modelling the interdependency of both users and items has proven beneficial \citep{dong2014knowledge,nickel2011three}. 
While GPLVM has been used as matrix factorization method for collaborative filtering in the past \citep{lawrence2009non}, as for other matrix decomposition approaches, practical applicability can be limited also for computational reasons. Being designed primarily as a dimensionality reduction method for data with few missing values, it is not readily amenable to the triple-based learning paradigm of collaborative filtering. That is, one trains a learning model with data samples in form of triples of indices and the corresponding target values, e.g. $(i, j, y_{i, j})$. The learning model will have to query the corresponding rows $i$ and $j$ from the respective factor matrices and attempt to regress them to the target entry $y_{i,j}$. 
This paradigm facilitates efficient modeling of large and highly sparse datasets such as large user-item matrices or even tensor representation of knowledge graphs \citep{nickel2015review}. However, the GPLVM-based recommender systems typically require storing the dense data matrix. More specifically, in the triple-based learning paradigm we only need iterate through all observed, i.e., non-zero scalar entries in a sparse matrix. %The GPLVM, would have to take as training label each entire row or column in the target matrix. 

In this work, we propose a novel approach to combine multi-output Gaussian processes with representation learning for collaborative filtering via matrix decomposition. 
We learn a latent representation vector for each entity type (i.e. users and items) jointly with a multi-output GP that predicts not only a point estimate of the user-item interaction, but also its predictive variance. We thereby explicitly model not only dependencies between users, but also between items.
We motivate this design by connections to coregionalization with separable kernels in multi-output Gaussian processes as well as to the GPLVM. We exploit variational sparse approximations \citep{titsias2009variational, hensman2015scalable} and combine them with a sparse matrix representation so that our approach scales to datasets consisting of large numbers of users and items via minibatch learning. 
Furthermore, we propose a new approach for evaluating and visualizing the quality of predictive variance of such uncertainty-aware models by stratifying predictions by their associated uncertainty.

\section{Related work} \label{sec:related_work}
Matrix decomposition methods applied to collaborative filtering and recommender systems attempt to learn explanatory latent features for users and items. 
Non negative matrix factorization \citep{luo2014efficient, zhang2006learning} restrict the factor matrices to be non-negative, in order to enable better interpretability of the representations.
Such methods are trained to minimize the reconstruction error in terms of the entire matrix, and are therefore no longer efficient or reliable in case of high sparsity in the matrix.
In contrast, biased matrix factorization \citep{koren2009matrix, gomez2015netflix} has proven to be very efficient in decomposing sparse matrices by training only based on observed entries. 
However, these methods are prone to overfitting and require careful tuning and validation \citep{nickel2015review}. 

The Gaussian Process Latent Variable Model (GPLVM) \citep{lawrence2004gaussian} applies GP regression in an unsupervised setting. 
Given a matrix $\bm{Y}$, the GPLVM learns a latent representation vector $\bm{a}_i$ for each row $\bm{y}_i$. 
A GP regression model is then used to map the unobserved latent variables $\bm{a}_i$ to the multiple output dimensions $\bm{y}_{\cdot, j}$. 
Since the latent variables are of lower dimensionality than the outputs, GPLVM can be seen as a dimension reduction approach. 
While GPLVM models are typically applied to dimensionality reduction tasks, \citep{lawrence2009non} apply GPLVM to large-scale matrices to perform item recommendation by collaborative filtering. While the GPLVM is usually viewed as a matrix factorization approach and has also an interpretation as probabilistic PCA,  it can also be interpreted as a multi-output GP regression model where the vectors of different outputs are drawn independently from the same Gaussian process prior.

Multi-output GP regression models can be used to predict vector-valued functions in a supervised manner by constructing a covariance matrix that describes the dependencies between all the input and output variables. Most implementations of multi-output GPs are formulated around the framework of the Linear Model of Coregionalization (LMC) \citep{alvarez2012kernels}, in the context of supervised learning. 
Within this framework, dependencies between individual regressors are modeled with separable kernels (or the sum thereof), which can be written as the product of two kernels. 
The first kernel measures similarity of samples in the input space, while the second kernel captures the similarity between each pair of output dimensions. 

\citep{dai2017efficient} extend multi-output GPs by representing the output dimensions with unobserved latent variables. 
Thereby they avoid problems with overfitting caused by having to estimate all parameters of the full covariance matrix in standard LMC, allow for the prediction of new outputs at test time and facilitate a lower computational complexity.
They further make sure their model scales to large datasets by utilizing the sparse Gaussian process approximation \citep{titsias2009variational, titsias2010bayesian} (SVGP). In the SVGP framework, the model is augmented by an auxiliary variable and the covariance matrix is computed on a set of inducing inputs that represent the whole dataset. 
The kernel distance between two input samples can then be factorized into their respective kernel distances to inducing points, thus avoiding the need to compute a full covariance matrix. %Another class of non-variational sparse approximation that involves inducing points is proposed in \citep{snelson2006sparse, snelson2007local}. 
%Here, inducing points are implemented and trained as kernel hyper parameters. 
For a systematic overview and comparison of sparse approximations, we refer to \citep{quinonero2005unifying} and \citep{bui2017unifying}.

\section{Preliminary: Matrix factorization for collaborative filtering} \label{sec:matrix_decompos}
In collaborative filtering, the dataset often takes the form of a user-item matrix $\bm{Y} \in \mathbb{R}^{I \times J}$, assuming $I$ users and $J$ items. 
Each entry $y_{i, j}$ describes the interaction between user $i$ and item $j$, for example in form of a rating or a purchase. 
Such a matrix is typically large and sparse, first due to the large number of users and items, and second due to the fact that each user is usually interested in a very small subset of items only. 
In order to derive new recommendations for a user, one has to predict their potential interest in an item with which no interaction has taken place yet, i.e. where there is no entry in the user-item matrix.
A typical solution to this collaborative filtering task is via a matrix decomposition approach.

During training, one fits a decomposition model that can recover all entries in the user-item matrix to a certain extent.
As a simple illustration, the user-item matrix can be modeled as $\bm{Y} \approx \bm{A}\bm{B}^T, ~\bm{A} \in \mathcal{A} = \mathbb{R}^{I \times r}, \bm{B} \in \mathcal{B} = \mathbb{R}^{J \times r}$, or equivalently, $y_{i, j} = \bm{a}_i^T\bm{b}_j ~\forall (i, j) \in [1, I]\times[1, J]$. 
At inference time, any entry value that is not observed in the training data, can be predicted by reading the corresponding location on the reconstructed the user-item matrix: $\widehat{\bm{Y}} = \bm{A}\bm{B}^T$. 
The generalization power of the decomposition model lies in the low-rank matrices $\bm{A}$ and $\bm{B}$, which can be interpreted as latent representations of the users and items, respectively. 
In order to achieve better modeling performance, it is possible to apply more complex functions than the dot product to join the latent representations. 
Writing the model definition as $y_{i, j} = g(\bm{a}_i, \bm{b}_j) ~\forall i, j \in [1, I] \times [1, J]$, the function $g$ can be realized in various ways. 
For instance, SVD defines
$g(\bm{a}_i, \bm{b}_j) = (\bm{\lambda} \circ \bm{a}_i)^T \bm{b}_j$, where $\bm{\lambda}$ consists of the singular values and $\bm{a}_i$ and $\bm{b}_j$ are restricted to be orthogonal.  
In case of biased matrix decomposition \citep{koren2009bellkor}, we have 
$g(\bm{a}_i, \bm{b}_j) = \bm{a}_i^T \bm{b}_j + \alpha_{i} + \beta_{j}$, where $\alpha_i$ and $\beta_j$ are user- and item-specific biases parameters, respectively. 
Multi-way neural networks \citep{dong2014knowledge}, if applied to the matrix case, take the form of
$g(\bm{a}_i, \bm{b}_j) = \bm{u}^T\sigma(\bm{w}^T[\bm{a}_i, \bm{b}_j]+w_0)+u_0$.
Here, the latent representations are concatenated and fed into a multi-layer perceptron.

The GPLVM \citep{lawrence2004gaussian, lawrence2007learning} solves the matrix decomposition task by fitting $J$ independent GP regression models on unobserved latent variables $\bm{x}_i$ with $p(\bm{y}_{\cdot,j})= \mathcal{N}(\bm{y}_{\cdot, j} ~|~ 0, \bm{K}^j+\sigma^2 \bm{I})$,
where $\bm{K}^j \in \mathbb{R}^{I \times I}$ captures the covariances between each pair of latent representation $(\bm{a}_i, \bm{a}_{i'})$ defined by the covariance function $k(\bm{a}_i, \bm{a}_{i'})$. In other words, the GP-LVM can be interpreted as a multiple-output GP regression model where only the output  data  are  given and the unobserved inputs are being optimized.

It is worth noting that the training target in case of standard GPLVM implementations is $\bm{y}_{\cdot, j}$. 
If the target matrix, and hence each column $\bm{y}_{\cdot, j}$, is sparse, such dense representations are computationally inefficient. 
In case of (biased) matrix decomposition and multi-way neural networks, however, one could formulate a training sample in form of a triple $(i, j, y_{i,j})$ and only iterate through the observed non-zero target values.

\section{Multi-output Gaussian Process Model for Collaborative Filtering}\label{sec:mwgp}
\subsection{Notation} \label{subsec:notation}
We denote a GP regression on target $\bm{y}$ as
\begin{align*}
    p(\bm{y} | \bm{f}) &= \mathcal{N}(\bm{y}~|~ \bm{f}, \sigma^2\bm{I}),
\end{align*}
where $\sigma$ is the hyper parameter that defines the random noise level. The vector $\bm{f}$ denotes the vector consisting of values produced by the regression model from all input data $\bm{x}_i$, which follows another multi-variate Gaussian distribution: 
\begin{align*}
    \bm{f} &= (
        f(\bm{x}_1), 
        f(\bm{x}_2), 
        ..., 
        f(\bm{x}_n) 
    )^T, \\
    ~ p(\bm{f}) &= \mathcal{N}(\bm{f} ~|~ \bm{0}, \bm{K}),  
\end{align*}
where we denote the covariance matrix using $\bm{K}$. 

\subsection{Model Definition} \label{subsec:model_def}
The GPLVM treats all output dimensions in the matrix as independent, which is an assumption that may not always hold, in particular if the matrix describes user-item interactions. For instance, the items in which a user takes interest could very well be correlated or similar \citep{nickel2015review, agrawal1994fast}. 
To capture such dependencies between output dimensions, we propose a new coregionalization kernel to perform multi-output Gaussian Process regression for unsupervised matrix decomposition in the spirit of the GPLVM.

Our proposed separable kernel can be written as the product of two individual kernels, where the first kernel measures similarity of samples in the input space and the second kernel captures the similarity between each pair of output dimensions. More formally, this kernel takes the form
\begin{align*} 
    (\bm{K})_{j, j'} &= k^A(\bm{a}_i, \bm{a}_{i'}) k^W(j, j')
\end{align*}
or, equivalently,
\begin{align*}
    \bm{K} &=  k^A(\bm{a}_i, \bm{a}_{i'})\bm{W} , %= \left( \bm{K(\bm{a}_i, \bm{a}_{i'})} \right)_{j, j'}.
\end{align*}
where $k^A$ and $k^W$ are scalar kernels on $\mathcal{A} \times \mathcal{A}$ and $\{1,\dots,J\} \times \{1,\dots,J\}$ respectively and $\bm{W}_{i, j} = k^W(i, j)$ is a symmetric and positive semi-definite matrix, which models the dependency between each pair of outputs. While coregionalization is usually performed in the context of supervised regression, in our application the inputs $\bm{a}_i$ are unobserved and, as for the GPLVM model, need to be optimized.
$\bm{W}$ being the identity matrix implies independence between outputs and the model falls back to the standard GPLVM. 
There is a variety of approaches for choosing $\bm{W}$, ranging from the design of a symmetric and positive semi-definite matrix based on suitable regularizers to the choice of covariance functions for the different output components \citep{alvarez2012kernels}.

Here, we choose to replace the coregionalization matrix $\bm{W}$ by a kernel on latent representations of items  in vector space $\mathcal{B}$, such that the covariance matrix can be written as a Kronecker product of the covariance matrix of the latent variables representing the items $\bm{K}^\mathcal{B}$ and the covariance matrix of the latent variables representing the users $\bm{K}^\mathcal{A}$.  We thereby combine representation learning with coregionalization in multi-output GPs for collaborative filtering via unsupervised matrix factorization.\\
Note that we have briefly introduced Latent Variable Multiple Output Gaussian Processes (LVMOGP) \citep{dai2017efficient} as a related method in Sec. \ref{sec:related_work}. 
In the context of supervised learning, LVMOGP augments multi-output GPs by latent variables, capturing latent information on the outputs. While the authors also replace $\bm{W}$ by a kernel on latent representations, in their supervised scenario, inputs are observed, whereas in our model both kernels act on latent variables. 
In Tab. \ref{tab:comparison} we compare the most closely related methods to ours, including multi-output GP regression, GPLVM and LVMOGP, in three aspects of learning task, output and input treatment, respectively. 

\begin{table*}[t]
    \centering
    \begin{tabular}{ r | c | c | c | c} 
    \hline 
                                        & Multi-output GPR  & GPLVM             & LVMOGP        & Ours  \\ \hline \hline 
         Unsupervised task              & $\times$          & $\checkmark$      & $\times$      & $\checkmark$ \\
         Dependency between outputs     & $\checkmark$      & $\times$          & $\checkmark$  & $\checkmark$ \\
         Coregionalization by kernel    & $\times$      & $\times$          & $\checkmark$  & $\checkmark$ \\
         Unobserved input               & $\times$          & $\checkmark$      & $\times$      & $\checkmark$ \\ \hline
    \end{tabular}
    \caption{A comparison of our methods with mostly related approaches. Coregionalization by kernel refers to learning latent vectors that represent output dimensions. In contrast, standard coregionalization handles dependency between outputs using a full covariance matrix whose only constraint is being symmetric and positive semi-definite.}
    \label{tab:comparison}
\end{table*}

Taken together, the kernel of our proposed model can be written as
\begin{align*}
    \bm{K}^{coreg} &= \bm{K}^\mathcal{A} \otimes \bm{K}^\mathcal{B} \\
    \text{with}~& \begin{cases}
    \bm{K}^\mathcal{A}_{i, i'} = k^\mathcal{A}(\bm{a_i}, \bm{a}_{i'}) = k_{\bm{A}}(i, i'), \\
    \bm{K}^\mathcal{B}_{j, j'} = k^\mathcal{B}(\bm{b_j}, \bm{b}_{j'}) = k_{\bm{B}}(j, j').     
    \end{cases}
\end{align*}
For the sake of symmetry, we choose the same covariance function for $k^\mathcal{A}$ and $k^\mathcal{B}$. 
As for matrix $\bm{K}^\mathcal{A}$, each element in $\bm{K}^\mathcal{B}$ measures the similarity between a pair of output dimensions (or items) $(j, j')$. 
As the rightmost term of the equations above implies, we could treat all latent variables in matrices $\bm{A}$ and $\bm{B}$ as hyper parameters in a specific kernel \citep{snelson2006sparse} that takes as input the indices. 
From a functional perspective, our proposed kernel 
\begin{align*}
    k^{coreg}((i, j), (i', j')) = k_{\bm{A}}(i, i') \cdot k_{\bm{B}}(j, j'),  
\end{align*}
measures in fact the similarity between user-item pairs $(i, j)$ and $(i', j')$. 
The measurement is carried out by evaluating the two kernel functions at $(i, i')$ and $(j, j')$, respectively, and calculating the product of both kernel values. 
The training samples can be thus formulated using only the indices and corresponding entries in the matrix: $(i, j, y_{i, j}) ~\forall i, j \in [1, I] \times [1, J]$.  
%\begin{align}\label{eq:triple}
%    (i, j, y_{i, j}) ~\forall i, j \in [1, I] \times [1, J]
%\end{align}
This data format is especially advantageous in case of large number of items and high sparsity in matrix $\bm{Y}$ since it does not require storing the data in dense matrix format.

We finally write our model as: $\bm{f} \sim \mathcal{N} (\bm{0}, \bm{K}^{coreg} + \sigma^2 \bm{I}) \in \mathbb{R}^{I \cdot J}, \bm{K}^{coreg} \in \mathbb{R}^{(I \cdot J) \times (I \cdot J)}$. 
The vector $\bm{f}$ has length $I \cdot J$, indicating that it consists of the outcome of all possible combinations of $I$ users and $J$ items.

\subsection{Model Fitting} \label{subsec:fitting}
The major challenge in fitting such a model lies in the size of the covariance matrix $\bm{K}^{coreg} \in \mathbb{R}^{(I \cdot J) \times (I \cdot J)}$, since there are typically large numbers of users and items. 
Computing the inverse of the covariance matrix, which is required for computing the log marginal likelihood, has a complexity of $\mathcal{O}(n^3)$ where $n$ is the number of training samples, in our case $n=I \cdot J$.
We take sparse GP approaches to address this challenge. 
A sparse GP model introduces $m \ll n$ inducing points, denoted as $\bm{Z}_m$ that represent the entire dataset. 
These inducing points are optimized as additional parameters in a GP model. 
%SPGP \citep{snelson2006sparse, snelson2007local} proposes to treat them as pseudo-inputs, i.e., parameters in the kernel definition, in a similar fashion that we define our index kernels $k_{\bm{A}}$ and $k_{\bm{B}}$. 
We apply SVGP \citep{titsias2009variational} and augment the model with inducing variables $\bm{u}$ with a Gaussian prior  $p(\bm{u})$,  that contain the values of function $f$ at inducing points $\bm{Z}_m$.
With this augmentation, a variational lower bound of the marginal log likelihood can be derived \citep{hensman2015scalable}: 
\begin{align} \label{eq:svgp_likelihood}
    \log p(\bm{y}) \geq &\log \mathcal{N}(\bm{y} | \bm{K}_{nm} \bm{K}_{mm}^{-1} \bm{\mu}, \sigma^2 \bm{I}) \nonumber \\
    &- \frac{1}{2\sigma^2} tr(\bm{K}_{nm}\bm{K}_{mm}^{-1}\bm{\Sigma}\bm{K}_{mm}^{-1}\bm{K}_{mn}) \\
    &- \frac{1}{2\sigma^2}tr(\bm{K}_{nn}-\bm{Q}_{nn}) - KL(q(\bm{u})||p(\bm{u})) \nonumber
\end{align}    
with $q(\bm{u})=\mathcal{N}(\bm{u}|\bm{\mu}, \bm{\Sigma})$ being the variational posterior distribution on the inducing variables $\bm{u}$, and $\bm{Q}_{nn} = \bm{K}_{nm}\bm{K}_{mm}^{-1} \bm{K}_{mn}$. $\bm{K}_{mm}$ is the covariance function evaluated between all the inducing points and $\bm{K}_{nm}$ is the covariance function between all  inducing points and training points. This formulation enjoys the advantage of being able to factorize across data, i.e., to be updated using single samples or mini batches, and can thus scale to large training data.  

The essential motivation of applying inducing points is to avoid calculating and inverting the full covariance matrix $\bm{K} \in \mathbb{R}^{n \times n}$.
Instead, one only calculates the kernel between a training sample and all inducing points as in $\bm{K}_{mn}$, and the kernel between two inducing points as in $\bm{K}_{mm}$, which is defined as follows in our specific case of coregionalization:  
\begin{align*} \label{eq:K_mm}
    \bm{K}_{mm} &= K(\bm{Z}^A, \bm{Z}^A) \otimes K(\bm{Z}^B, \bm{Z}^B) \\
    &\in \mathbb{R}^{(m_A m_B) \times (m_A m_B)}.
\end{align*}

SVGP therefore reduces the computational complexity from 
$\mathcal{O}((I \cdot J)^3)$ to $\mathcal{O}((I \cdot J) \cdot (m_A \cdot m_B)^2)$. Our triple-based formulation means that the set of training samples consists only of observed triples; by making use of the SVGP approximation, we only ever need to compute the kernel between such a training sample and all inducing points or between pairs of inducing points. This in turn means that for matrices with sparsity $S$ (fraction of \emph{observed} user-item interactions), the computational complexity reduces to  $\mathcal{O}(S \cdot(I \cdot J) \cdot (m_A \cdot m_B)^2)$. 
In practice, we can further reduce this computational complexity if we choose the same number of inducing points $m$ in $\mathcal{A}$ and $\mathcal{B}$: this allows us to learn coupled pairs of inducing points in both spaces (see next subsection for implementation details). By tying the parameters of both sets of latent representations, we can then reduce the effective size of $ \bm{K}_{mm}$ to $m \times m$ and further reduce computational complexity to  $\mathcal{O}(S\cdot(I \cdot J) \cdot m^2)$. 

%In summary, our proposed model performs a sparse matrix decomposition, where the representation vectors of users and items are consumed by a GP regression with specifically designed kernel.
%The kernel measures the similarity between two user-item pairs in form of $(i, j)$ and $(i', j')$
%At inference, the model generates a Gaussian distribution whose mean serves as point estimate and the variance as prediction uncertainty. 

\subsection{Coupling Inducing Points} \label{subsec:coupling_inducing_points}
%Many matrix and tensor decomposition methods for collaborative filtering aim at modeling interactions between entities in each dimension \citep{nickel2015review}.
The adjacency matrix characterizing interactions between users and items in collaborative filtering tasks is typically highly sparse. An efficient representation of such data is via a triple store where an observed training sample is represented as $(i, j, y_{i, j}) ~\forall (i, j) \in [1, I] \times [1, J]$ with users $i$, items $j$ and corresponding entries in the user-item matrix $y_{i,j}$. In this data formulation for sparse data, input $i$ and output $j$ of one training sample are inherently coupled. We would like for this coupling to be also reflected in the sparse approximation and in particular the choice of inducing points, in order to fully take advantage of the benefits of this data representation in terms of computational complexity.\\
In standard formulations of SVGP for multi-output GPs, the covariance matrix of the inducing variables $\bm{K}_{mm}$, is computed as $\bm{K}_{mm}=\bm{K}^A_{mm} \otimes \bm{K}^B_{mm}$. Here, $\bm{K}^A_{mm}=K^A(\bm{Z}^A, \bm{Z}^A)$ and $\bm{K}^B_{mm}=K^B(\bm{Z}^B, \bm{Z}^B)$ are computed on independent sets of inducing points $\bm{Z}^A$ and $\bm{Z}^B$ with kernel functions $k_A$ and $k_B$ that represent input space $\mathcal{A}$ and output space $\mathcal{B}$ respectively \citep{dai2017efficient}. Similarly, the cross-covariance between training samples and inducing points $\bm{K}_{nm}$ is constructed as $\bm{K}_{nm}=\bm{K}^A_{nm} \otimes \bm{K}^B_{nm}$,  with $\bm{K}^A_{nm} \in \mathbb{R}^{I \times m_A}$ and $\bm{K}^B_{nm} \in \mathbb{R}^{J \times m_B}$, where $I$ and $J$ correspond to the number of inputs (users) and outputs (items) as before. $\bm{K}^A_{nm}$ is the cross-covariance between latent inputs $\bm{A}$ and $\bm{Z}^A$ and $\bm{K}^B_{nm}$ the cross-covariance between latent outputs $\bm{B}$ and $\bm{Z}^B$. The covariance between training samples $\bm{K}_{nn}$ is computed as $\bm{K}_{nn} = \bm{K}^A_{nn} \otimes \bm{K}^B_{nn}$, with $\bm{K}^A_{nn}$ being the covariance matrix of $\bm{A}$ constructed with $k_A$ and $\bm{K}^A_{nn}$ being the covariance matrix of $\bm{B}$ constructed with $k_B$.\\
However, this formulation does not reflect the sparse triple-based data formulation and results in a computational complexity that is quartic in $m$ (c.f. previous section). If we decide to choose the same number of inducing inputs for $\bm{Z}_A$ and $\bm{Z}_B$, we can couple the inducing points for pairs of inputs and outputs and reformulate the construction of $\bm{K}_{nm}$, $\bm{K}_{nm}$ and $\bm{K}_{nn}$, such that it reflects the coupling between input $i$ and output $j$ of a training sample $(i,j)$.\\ More specifically, when we compute the cross-covariance between the $q$-th training sample $(i, j)$ \footnote{There is always a bijection / double-indexing relation between $(i, j)$ and $q$.} and the $l$-th pair of inducing points in order to build $\bm{K}_{nm}$, we do so by pairing the corresponding latent representations $\bm{a}_i$ and $\bm{b}_j$ and using a product kernel to compute similarities. That is, 
\begin{equation}
    \begin{aligned} \label{eq:K_nm}
    & K_{nm}((\bm{A}, \bm{B}),(\bm{Z}^A, \bm{Z}^B))_{q,l} \\
    =& k([\bm{a}_i,\bm{b}_j], [\bm{z}^A_l, \bm{z}^B_l]) \\
    =& k_A(\bm{a}_i, \bm{z}^A_l) k_B(\bm{b}_j, \bm{z}^B_l)
    \end{aligned}
\end{equation}
Using this paired formulation, it becomes immediately clear that \[\bm{K}_{nm} = K_{nm}((\bm{A}, \bm{B}),(\bm{Z}^A, \bm{Z}^B))\] becomes substantially smaller,  with $\bm{K}^{\mathrm{paired}}_{nm} \in \mathbb{R}^{(I\cdot J)\times m}$ compared to the standard case where the sets of inducing points are treated as independent and $\bm{K}_{nm} \in \mathbb{R}^{(I \cdot J) \times (m_A \cdot m_B) }$.\\
We construct the covariance matrix $\bm{K}_{mm}$ using the same pairing approach and compute the covariance between the $o$-th and $p$-th inducing points as 
\begin{equation}
\begin{aligned} \label{eq:K_mm}
&K_{mm}((\bm{Z}^A, \bm{Z}^B),(\bm{Z}^A, \bm{Z}^B))_{o,p} \\
=& k([\bm{z^A}_o,\bm{z^B}_o], [\bm{z^A}_p, \bm{z^B}_p]) \\
=& k_A(\bm{z^A}_o, \bm{z^A}_p) k_B(\bm{z^B}_o, \bm{z^B}_p)
\end{aligned}
\end{equation}
That means, using paired inducing points, $\bm{K}_{mm}$ can  be written as elementwise product between $\bm{K}^A_{mm}$ and $\bm{K}^B_{mm}$:  $\bm{K}_{mm} = \bm{K}^A_{mm} \odot \bm{K}^B_{mm}$. As before, $\bm{K}_{mm}$ becomes substantially smaller since it is constructed via an elementwise product rather than via the Kronecker product.\\
Finally, we use the same pairing formalism when constructing $\bm{K}_{nn}$ and compute the covariance between the $q$-th training sample $(i,j)$ and the $l$-th training sample $(i', j')$ as
\begin{align*}
& K_{nn}((\bm{A}, \bm{B}),(\bm{A}, \bm{B}))_{k,l} \\
=& k([\bm{a}_i,\bm{b}_j], [\bm{a}_{i'}, \bm{b}_{j'}]) \\
=& k_A(\bm{a}_i, \bm{a}_{i'}) k_B(\bm{b}_j, \bm{b}_{j'})
\end{align*}
Constructing $\bm{K}_{nn}$ in this manner rather than using the Kronecker product between $\bm{K}^A_{nn}$ and $\bm{K}^B_{nn}$ is particularly suited for highly sparse data, where the number of training samples $N$ is substantially lower than total number of possible samples $I \cdot J$. In this case the paired approach directly leads to a natural construction of a small covariance matrix of observed training samples of size $N\times N$\footnote{Note: the full covariance matrix is not needed during inference. Model training using SVGP merely requires the computation of the trace of $\bm{K}_{nn}$}.\\
Taken together, this means that the paired approach reduces the computational complexity of  of our method from $\mathcal{O}(I \cdot J \cdot m_A^2 \cdot m_B^2)$ to $\mathcal{O}(S\cdot(I \cdot J) \cdot m^2)$  (with sparsity $S$, c.f. section \ref{subsec:model_def}).

We summarize the training algorithm of our approach as follows. 
Given the number of inducing pairs $m$ and rank $r$ as hyper parameters:  
\begin{algorithm}
\SetAlgoLined
\SetKwInOut{Input}{input}
\SetKwInOut{Output}{output}
\Input{Training $N$ tuples $(i, j, y_{i,j})$ where $i \in [1, I], j\in [1, J]$, and $y_{i, j}$ is the $(i, j)$-th entry in matrix $\bm{Y}$;} 
\Output{$\bm{A}, \bm{B}, \bm{Z}^A, \bm{Z}^B, \bm{\mu}, \bm{S}$ and the RBF parameters;} 
Initialize latent representations $\bm{A} \in \mathbb{R}^{I \times r}, \bm{B} \in \mathbb{R}^{J \times r}$, and inducing pairs $\bm{Z}^A \in \mathbb{R}^{m \times r}, \bm{Z}^B \in \mathbb{R}^{m \times r}$;\\ 

\While{not converged}{
\For{each training epoch}{
  \For{a training sample $(i, j, y_{i, j})$}{
    Calculate the covariance matrix of inducing pairs $\bm{K}_{mm}$ using Eq. \eqref{eq:K_mm}; \\
    Derive the bijective index $q$ of $(i, j)$; \\
    Calculate its covariance values $K_{nm}[q, l]$ with all inducing pairs, using Eq. \eqref{eq:K_nm}. 
    Update the ELBO in Eq. \eqref{eq:svgp_likelihood} w.r.t. $\bm{a}_i, \bm{b}_j, \bm{z}^A_l, \bm{z}^B_l, \bm{\mu}, \bm{S}$ and the RBF kernel parameters. 
  }
 }
 }
 \caption{Training algorithm}
\end{algorithm}
We describe our algorithm as online training with single samples for a more concise notation as well as to emphasize the scalability of the training. 
In the step of updating the ELBO, one only needs to iterate once through all inducing pairs. In practice, one would often like to train on batches. The complexity of this case merely becomes the product of batch size and number of inducing pairs. 
One should also note that in a standard SVGP setting, one updates the kernel parameters and inducing points. In our approach, however, we update additionally the \emph{input} to the GP which are the latent representations. 

We have implemented our method using tensorflow/GPflow \citep{de2017gpflow} and provide an easy-to-use python package at \url{https://github.com/Tuyki/mogp-decomposition} with experimental results.  
For all experiments, we used an RBF kernel, for both $k_A$ and $k_B$. Experiments were run on hardware utilising 4 CPUs and one GeForce GTX TITAN X GPU.

\subsection{Evaluation} \label{subsec:evaluation}
Due to the fact that the missing entries in CF tasks may be not be completely random, high performance in terms of MAE and RMSE might not always guarantee good recommendations \citep{marlin2012collaborative, cremonesi2010performance}. 
To this end, we propose an additional metrics to evaluate uncertainty-aware CF models, including ours and GPLVM, by sorting the point estimates based on the predictive variance. 
Specifically, we denote the ground truth values as $(y_i)_{i=1}^{n}$, and the predictive distribution as $(f_i \sim \mathcal{N}(m_i, s^2_i))_{i=1}^{n}$. 
We calculate $Q$ quantiles of all the standard deviations $(s_i)_{i=1}^{n}$. % and denote these as $(q_u)_{u=1}^{Q}$. 
For each quantile $q \in \{\frac{1}{Q}, \frac{2}{Q}, \dots, 1\}$ where the predictive standard deviation $s_i$ is smaller than or equal to the $q$-quantile, we evaluate all point estimates $m_i$ against the corresponding ground truth values $y_i$. 
Formally let, 
\begin{align*}
    d_q = R(\{ (y_i, m_i) ~|~ \forall s_i \leq q\text{-Quantile} \}),
\end{align*}
where $R$ could be any evaluation metric defined on a set of pairs of ground truth and point estimate. 
Each $d_q$ is the evaluation of the predictive performance of the model on its top-$q \cdot n$ most confident predictions. 
Ideally, an uncertainty-aware prediction model is expected to have better performances on test samples about which it demonstrates higher confidence, i.e., lower predictive variance, and vice versa. 
If we plot all $d_q$ values on y-axis against the corresponding $q$ values on the x-axis, we would expect a monotonously increasing plot line. 
We name this evaluation approach QP-plot, where Q stands for quantile and P for performance.

\section{Experiments}\label{sec:experiments}

We conduct experiments on three well established datasets for evaluating recommender systems: movielens-1m (ML-1M), movielens-10m (ML-10M) and Jester joke recommendations. 
The task is to predict the rating value of an arbitrary pair of user and item, which can be later interpreted as a score quantifying the interest of the user in the item. 

The ML-1M consists of 1000209 ratings of 6040 users on 3706 items, the ML-10M consists of 10000054 ratings of 69878 users on 10677 items, and the jester dataset consists of 1728847 ratings of 50692 users on 140 items. 
The ML-1M has as ratings only natural numbers between 1 and 5, while ML-10M has rating in the same range but the ratings also include one value between each pair of natural numbers, such as $\{1, 1.5, 2, 2.5, ..., 5\}$.
The Jester dataset contains arbitrary real numbers between -10 and 10. 
Our model, based on a GP regression, models all three types of ratings as real values.

We performed a hyper-parameter search especially with regard to the size of the latent vectors and number of inducing points. 
%We realize that the model learns more efficiently with larger mini batches. 
On a grid search of batch sizes $\{ 2^8, 2^{10}, 2^{12}, 2^{14}, 2^{16}, 2^{18}\}$, a relatively large size of $2^{16}$ produced the best performance.  
We also perform a grid search for the optimal number of inducing points. 
Among $\{32, 64, 128, 256, 512, 1024\}$ inducing points, $128$ outperforms other choices. 
This comparison is also illustrated as QP-plot in supplementary materials.  
We also find out that the model is robust with respect to the size of latent vector. 
For all the experiments, we use the best performing size of 8, which is a relatively small number compared to, e.g., SMA \citep{li2016low}, which requires a rank of up to $50$. 

For each dataset we report the performance in terms of MAE and RMSE on the test set in a 5-fold cross validation setting. 
Quantitative comparisons regarding the ML-1M, ML-10M and Jester datasets can be found in Tab. \ref{tab:ml-1m}, \ref{tab:ml-10m} and \ref{tab:jester}, respectively.  
In each table, We have two groups of evaluation references. 
First, we refer to the leading results that are reported by other works on the same dataset, but not necessarily with the same cross validation split.
When applicable, we include the reported ranks of these methods.  
Second, we conduct new experiments with Biased MF, SVD++ and Bayesian GPLVM \citep{titsias2010bayesian} using exactly the same cross validation setting and the same size of latent vectors. 
We show the average performance of our model on a varying standard deviation quantile between 80\% and 100\%, and highlight (in shade) the values between which the best reported performances fall.

\begin{table}[t]
    \centering
    \begin{tabular}{ l | c | c } \hline
                                        & RMSE    & MAE     \\ 
                                        \hline \hline 
         CF-NADE \citep{zheng2016neural}         & 0.829   & -       \\
         GPLVM \citep{lawrence2009non}         & 0.880     & \textbf{0.644}   \\
         Sparse FC \citep{muller2018kernelized}    & \textbf{0.824}   & -       \\
         GC \citep{berg2017graph}           & 0.832   & -       \\
         CWOCF \citep{lu2013second}            & 0.958   & 0.761   \\        
         LLORMA \citep{lee2013local}            & 0.865   & -       \\
         \hline \hline 
         Biased MF                      & 0.863   & 0.678   \\
         SVD++                          & 0.893   & 0.705   \\
         Bayesian GPLVM at $q=100\%$                & 0.889   & 0.698  \\ 
         \hline \hline 
         MW-GP at $q=100\%$               & 0.866   & 0.676   \\
         MW-GP at $q=90\%$               &\cellcolor{gray!25}0.838   &\cellcolor{gray!25}0.657   \\
         MW-GP at $q=80\%$               &\cellcolor{gray!25}0.821   &\cellcolor{gray!25}0.643   \\
         \hline
    \end{tabular}
    \caption{Performance comparison on the ML-1M dataset.}
    \label{tab:ml-1m}
\end{table}

\begin{table}[t]
    \centering
    \begin{tabular}{ l | c | c } \hline 
                                        & RMSE   & MAE \\ \hline \hline 
         CF-NADE \citep{zheng2016neural}         & 0.771  & -         \\
         GPLVM \citep{lawrence2009non}         & 0.874 & 0.635         \\
         Sparse FC\citep{muller2018kernelized}    & 0.769  & -         \\
         GC \citep{berg2017graph}           & 0.777  & -      \\
         CWOCF \citep{lu2013second}            & 0.903  & 0.701 \\
         LLORMA \citep{lee2013local}            & 0.822  & - \\
         SMA \citep{li2016low}               & \textbf{0.768}  & - \\ 
         \hline \hline
         Biased MF                      & 0.797  & \textbf{0.612} \\
         SVD++                          & 0.826  & 0.636 \\ \hline \hline
         MW-GP at $q=100\%$             & 0.808  & \cellcolor{gray!25}0.618 \\
         MW-GP at $q=90\%$             & \cellcolor{gray!25}0.774  & \cellcolor{gray!25}0.594 \\
         MW-GP at $q=80\%$             & \cellcolor{gray!25}0.753  & 0.579 \\
         \hline
    \end{tabular}
    \caption{Performance comparison on the ML-10M dataset.}
    \label{tab:ml-10m}
\end{table}

\begin{table}[t]
    \centering
    \begin{tabular}{ c | c | c } \hline 
                                        & RMSE   & MAE \\ \hline
        \citep{desrosiers2010novel}      & 4.480  & 3.541 \\
        \citep{goldberg2001eigentaste}   & -      & 3.740 \\ \hline \hline 
         Biased MF                      & 5.826  & 4.300 \\
         SVD++                          & 4.253  & 3.117  \\ \hline 
         MW-GP at $100\%$               & \textbf{4.171}  & \textbf{3.092} \\
         MW-GP at $90\%$               & 4.034  & 3.019 \\ 
         MW-GP at $80\%$               & 3.916  & 2.944 \\
         \hline
    \end{tabular}
    \caption{Performance comparison on the Jester dataset.}
    \label{tab:jester}
\end{table}

\begin{figure}[ht!]
    \includegraphics[scale=0.32, trim={0 0 0 0}, clip]{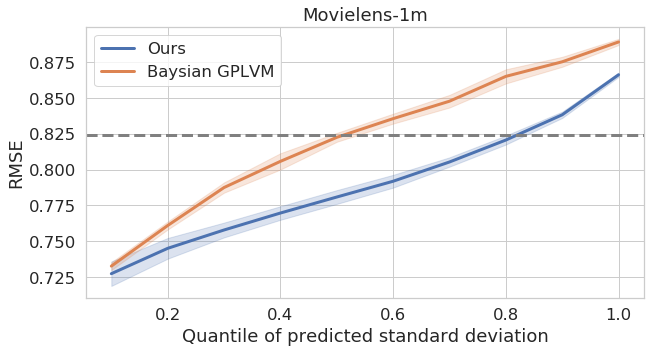}
    \includegraphics[scale=0.32, trim={0 0 0 0}, clip]{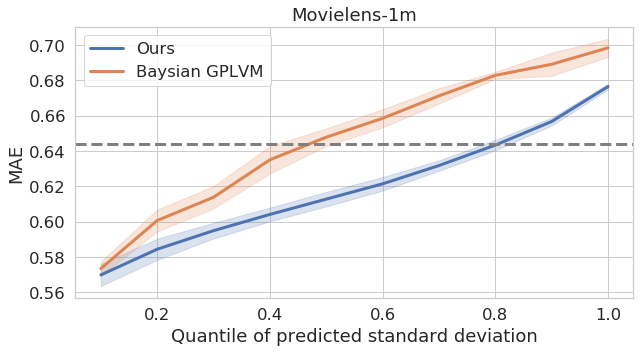}
    \caption{QP plots regarding the ML-1M dataset, comparing our method with Bayesian GPLVM and best reported performances. The horizontal dashed lines correspond to reported state-of-the-art methods. RMSE: \citep{li2016low}, MAE: \citep{lawrence2009non}.}
    \label{fig:res_ml1m}
    \hspace{4cm}
\end{figure}

\begin{figure}[ht!]
    \includegraphics[scale=0.32, trim={0 0 0 0}, clip]{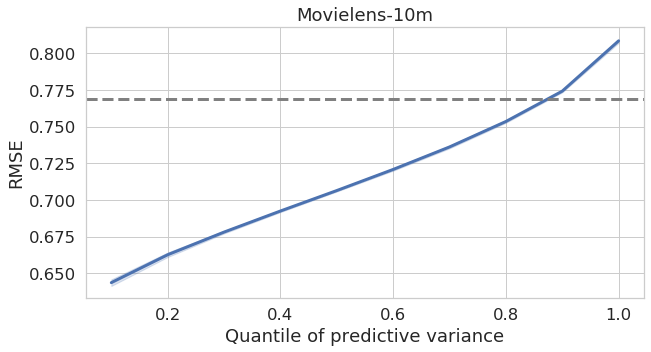}
    \includegraphics[scale=0.32, trim={0 0 0 0}, clip]{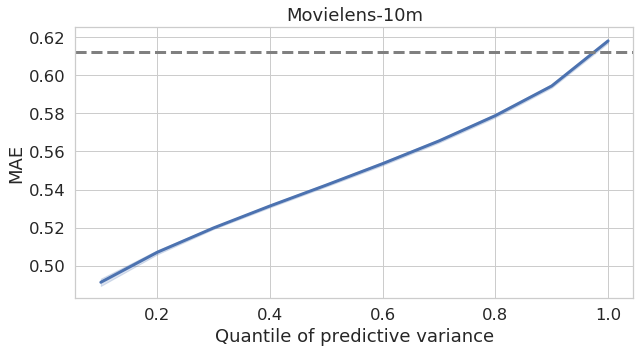}
    \caption{QP plots regarding the ML-10M dataset, comparing our method with Bayesian GPLVM and best reported performances. The horizontal dashed lines correspond to reported state-of-the-art methods. RMSE: \citep{zheng2016neural}. In terms of MAE, the Biased MF outperforms our method only at quantile 100\%. The GPLVM seems to suffer from the data volume and fails to converge in training.}
    \label{fig:res_ml10m}
    \hspace{4cm}
\end{figure}

\begin{figure}[ht!]
    \includegraphics[scale=0.32, trim={0 0 0 0}, clip]{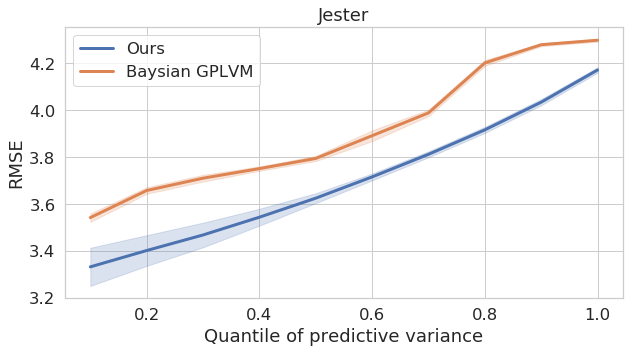} 
    \includegraphics[scale=0.32, trim={0 0 0 0}, clip]{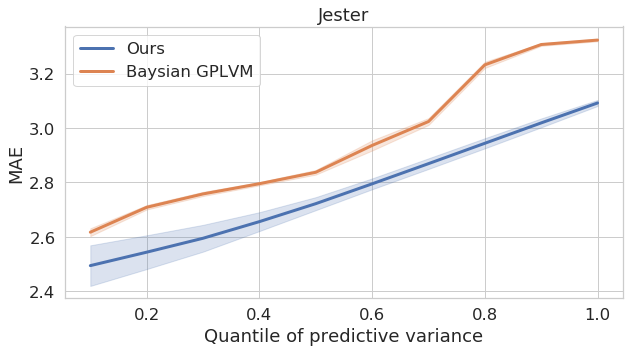}
    \caption{QP plots regarding the Jester dataset, comparing our method with Bayesian GPLVM. We do not find reported evaluations that outperform ours. }
    \label{fig:res_jester}
\end{figure}

We also demonstrate performances of our method using the QP-plot in Fig. \ref{fig:res_ml1m}, \ref{fig:res_ml10m} and \ref{fig:res_jester}. 
We visualize the average performance of all 5 splits with a solid line and the standard deviation across the 5 splits as error-bars. 
On the plots regarding the ML-1M and ML-10M datasets, we plot as dashed horizontal line the best performance reported in literature, which cut our curve between the quantiles of 80\% and 90\% in both of ML-1M in terms of MAE and RMSE. 
In the case of ML-10M, the best reported RMSE lies between our quantile 90\% and 100\%, the best reported MAE between 90\% and 100\%. 
In other words, for those $70\%$ to $90\%$ of test samples for which our model is most confident, it also produces equal or better point estimates than state-of-the-art recommender systems. It is also worth noting that our model outperforms the only uncertainty-aware model, GPLVM, in all quantiles of variances, presumably due to the explicit modeling of output dependency.

Our training time on the ml-1m dataset is 13.4 min (1 CPU, 1 GPU) for all 500 epochs. For the same setting, the biased MF model requires 16.1 min and SVD++ 290.8 min (1 CPU). The uncertainty-aware Bayesian GPLVM has a substantially longer training time of more than 10 hours (16 CPUs, wall time).
We perform inference for all tuples in the test set simultaneously. In case of ml-1m, for a test set of size ~200K, the average inference time is 0.506 seconds. This demonstrates the computational efficiency of our solution to couple the inducing points (section \ref{subsec:coupling_inducing_points}). We also realize that, since the kernel evaluation is independent for different pairs of vectors, our model can profit from the parallelization power of GPUs.

\section{Conclusion and Discussion} \label{sec:discussion}
In this manuscript, we propose a novel uncertainty-aware approach to perform collaborative filtering in the context of recommender systems. To this end, we introduce a coregionalization kernel to perform multi-output Gaussian Process regression for unsupervised matrix decomposition in the spirit of GPLVM. In contrast to the GPLVM we do not treat output dimensions as independent, but model a coregionalization matrix with a kernel on latent variables representing the output.
Our approach not only accounts for dependencies between output dimensions, but also allows our model to handle extremely sparse data in form of triple-stores in an efficient manner. 

Finally, we would like to stress that the aim of our method is not to improve prediction quality that can be measured by metrics such as RMSE or MAE. Instead, we are interested in exploring the possible solutions to add uncertainty awareness to collaborative filtering approaches and recommender systems. 
More specifically, we enable these methods to express its uncertainty about every prediction.  
With the help of our evaluation method, the QP-plot, we can show that such uncertainty quantification are valid: for predictions that deviate from the ground truth, the corresponding predictive variances tend to be larger. 
With our experiments, we also show that Bayesian GPLVM fails to scale to large datasets in terms of both training and inference. Our approach in contrast merely requires training and inference time that is comparable to highly optimized linear models such as SVD++. 

\section{Acknowledgment}
The authors thank Markus Kaiser for his valuable insights and discussions on variational sparse Gaussian Processes. 

\bibliography{yang_573}

\end{document}